\documentclass[conference]{IEEEtran}
\IEEEoverridecommandlockouts
\usepackage{cite}
\usepackage{amsmath,amssymb,amsfonts}
\usepackage{algorithmic}
\usepackage{graphicx}
\usepackage{textcomp}
\usepackage{xcolor}
\usepackage{booktabs}
\usepackage{makecell}
\usepackage[linesnumbered,ruled,vlined]{algorithm2e}
\SetKwInput{KwInput}{Input}
\SetKwInput{KwOutput}

\usepackage{url}
\usepackage{tikz}
\usetikzlibrary{arrows.meta, positioning, shapes.geometric, backgrounds, shapes.symbols, fit}

\def\BibTeX{{\rm B\kern-.05em{\sc i\kern-.025em b}\kern-.08em
    T\kern-.1667em\lower.7ex\hbox{E}\kern-.125emX}}
\begin{document}

\title{Compressed Orientation Reachability Maps for Sub-Second Robotic Needle Entry Point Planning\\
}

\author{
    \IEEEauthorblockN{
        Prabhu Nithin Gollapudi\IEEEauthorrefmark{1},
        Andreas Maier\IEEEauthorrefmark{1},
        Franziska Mathis-Ullrich\IEEEauthorrefmark{1},
        Water Zhao\IEEEauthorrefmark{2},
        Jia Wang\IEEEauthorrefmark{2}, and
        Siming Bayer\IEEEauthorrefmark{1}
    }
    % \vspace{0.5ex} % Optional: adds a tiny bit of breathing room before affiliations
    \IEEEauthorblockA{
        \IEEEauthorrefmark{1} Friedrich-Alexander-Universität Erlangen-Nürnberg, Erlangen, Germany\\
        Email: \{prabhu.nithin.gollapudi, andreas.maier, franziska.mathis-ullrich, siming.bayer\}@fau.de
    }
    % \vspace{1ex}
    % \IEEEauthorblockA{
    %     \IEEEauthorrefmark{2}Surgical Planning and Robotic Cognition Lab, 
    %     Friedrich-Alexander-Universität Erlangen-Nürnberg, Erlangen, Germany\\
    %     Email: franziska.mathis-ullrich@fau.de
    % }
    % \vspace{0.5ex}
    \IEEEauthorblockA{
        \IEEEauthorrefmark{2}Precision Medtech Co., Ltd, Beijing, China\\
        Email: \{water.zhao, jia.wang\}@precisionmedtech.com
    }
}

\maketitle

\begin{abstract}
% The abstract should be a single paragraph, 150-250 words.
% Context -> Problem -> Proposed Method -> Results -> Impact.

In the context of multi-stage trajectory planning for robotic percutaneous liver interventions, identifying a kinematically feasible entry point remains a critical computational bottleneck. Existing methods rely on exhaustive inverse kinematics or randomized searches, incurring delays of 13 to 27 seconds per target and limiting real-time intraoperative adaptation. This paper introduces an Orientation-Based Reachability Map (ORM) that precomputes the kinematic feasibility of a 6 Degrees of Freedom (DoF) serial manipulator across 188,873 workspace positions and encodes orientation feasibility into a compact, indexed 83\,MB map. A Just-In-Time (JIT) compiled runtime pipeline replaces live optimization with near-instantaneous array lookups. Experimental evaluation demonstrates a reduction in planning latency from 13.5--26.9\,s to 0.289--0.349\,s per target while maintaining high geometric fidelity, achieving a mean convex hull volume overlap ratio of 0.933 and a mean alpha-shape volume ratio of 0.976 relative to the baseline reachable point cloud. The proposed ORM framework enables sub-second entry point planning and provides an efficient computational framework for robotic needle insertion in percutaneous liver ablation.

\end{abstract}


\section{Introduction}

Liver cancer remains one of the leading causes of cancer-related mortality worldwide, with approximately 865,000 new cases and 757,948 deaths reported in 2022 \cite{globocan}, and a 5-year survival rate of only 22\% \cite{uscancer}. For patients with underlying liver disease who are not eligible for surgical resection, percutaneous thermal ablation techniques such as radiofrequency ablation (RFA) and microwave ablation (MWA) are established first-line therapies for small hepatocellular carcinoma (HCC) and selected liver metastases \cite{liang, uribe, tinguely}. These minimally invasive procedures offer reduced morbidity, shorter hospital stays, and tissue preservation, but their effectiveness depends critically on accurate needle placement to ensure complete tumor coverage with an adequate safety margin (typically 5--10\,mm) while avoiding major vessels and bile ducts  \cite{tinguely, beyer, Schaible2020-go, bodard}.

Robotic assistance offers clear advantages over freehand techniques by reducing tremor, operator variability, and radiation exposure \cite{uribe, Schaible2020-go, bodard, beyer}. In image-guided ablation, treatment success depends on a multi-stage planning pipeline that includes tumor targeting, ablation volume simulation, and needle trajectory selection \cite{Li2025, Pan2026}. Within this pipeline, identifying a kinematically feasible skin entry point is a necessary prerequisite, since even a clinically optimal anatomical trajectory cannot be executed if the manipulator cannot reach it \cite{Ding2026-gf, Li2025}.

However, existing intraoperative entry point planning methods often rely on exhaustive inverse kinematics (IK) verification or randomized search to evaluate manipulator reachability over large candidate sets. This leads to latencies of 13--27\,s per target \cite{Pan2026, uribe}, which is slow for responsive intraoperative replanning under respiratory motion and soft-tissue deformation.

To address this limitation, we propose an Orientation-Based Reachability Map (ORM) framework for fast kinematic feasibility evaluation in robotic needle guidance. The specific contributions of this paper are:

\begin{itemize}
\item An ORM representation that precomputes reachability for a 6 Degrees of Freedom (DoF) serial manipulator over 188,873 workspace positions, storing orientation feasibility in a compact 83\,MB indexed structure.
\item A Just-In-Time (JIT) compiled query pipeline that replaces repeated online inverse kinematics (IK) evaluations with indexed lookups.
\item An experimental evaluation demonstrating a reduction in entry point planning latency from 13.5--26.9\,s to 0.289--0.349\,s per target (a 60--90$\times$ speedup), while preserving high geometric fidelity with a mean alpha-shape volume ratio of 0.976 and a mean convex hull volume overlap ratio of 0.933.
\end{itemize}

These results support the use of the proposed ORM as a practical foundation for sub-second, workspace-aware entry point planning in robotic percutaneous liver ablation.

The remainder of this paper is organized as follows. Section II reviews related work in reachability analysis and robotic surgery. Section III details the proposed ORM generation and runtime pipeline. Section IV presents the experimental setup and results. Finally, Section V concludes the paper and discusses future work.

\section{Related Work}

\subsection{Path and Entry Point Planning in Medical Robotics}

A substantial body of research has addressed needle trajectory planning for percutaneous interventions within the broader scope of image-guided ablation. These holistic, multi-stage approaches primarily optimize clinical criteria such as insertion depth, avoidance of critical structures including major vessels and bile ducts, and sufficient treatment margins around the target lesion \cite{Li2025, LUO2022105506}. While highly effective for anatomical path selection, many of these methods treat robot execution as a decoupled problem. By deferring manipulator reachability validation to a later, secondary stage, the overall pipeline suffers from increased computational costs and limited intraoperative responsiveness \cite{Ding2026-gf, Pan2026}. Therefore, resolving the entry point feasibility bottleneck is essential to realizing the full potential of these comprehensive planning methods.

\subsection{Kinematic Feasibility and Online Planning}

Robotic execution of a planned entry point is commonly verified using IK or sampling-based planning methods. Jacobian-based numerical IK solvers and search-based approaches can provide robust feasibility assessment, but their computational cost becomes significant when dense entry points must be evaluated online \cite{Belbachir, sundaram2022task}. In percutaneous liver ablation, this cost is especially problematic because respiratory motion and soft-tissue deformation require fast replanning, whereas exhaustive online kinematic validation can require 13--27\,s per target \cite{Pan2026}.

\subsection{Reachability Maps for Fast Feasibility Queries}

To reduce the burden of repeated online IK evaluation, reachability maps have been proposed as precomputed workspace representations for fast feasibility lookup. Porges et al. introduced capability maps as discrete spatial models for rapid query-based assessment of manipulator feasibility \cite{porges-main}. In minimally invasive robotic surgery, related workspace-analysis methods have been used for base placement, trocar positioning, and system setup optimization \cite{reuleaux, junorm, Lohmann2012AWA, 9665151}. However, these methods are largely aimed at coarse setup decisions or positional reachability only, and extending them to dense 6-DoF needle insertion queries can lead to substantial discretization error or impractical memory demands \cite{reuleaux, junorm, 7353564}.

\subsection{Research Gap}

Despite progress in anatomical planning, kinematic validation, and workspace precomputation, an important gap remains for dense intraoperative needle insertion planning. Existing clinical planners do not provide sufficiently fast integrated kinematic verification, while current reachability-based methods are not well suited for high-resolution 6-DoF query evaluation under surgical runtime constraints. Our work addresses this gap by introducing a memory-efficient ORM designed for sub-second, workspace-aware entry point evaluation.

\section{Methodology}

\begin{figure}[htbp] 
\centering
% \textwidth forces the entire 2-column diagram to fit perfectly inside your page margins
\resizebox{\columnwidth}{!}{%
\begin{tikzpicture}[
    >=Stealth,
    node distance=0.8cm and 1.2cm,
    shared/.style={rectangle, draw=black, thick, fill=purple!10, text width=5cm, align=center, rounded corners, minimum height=1cm, font=\small\bfseries},
    block/.style={rectangle, draw=black, thick, fill=blue!5, text width=3.5cm, align=center, rounded corners, minimum height=1.2cm, font=\small},
    bottleneck/.style={rectangle, draw=red, thick, fill=red!5, text width=3.5cm, align=center, rounded corners, minimum height=1.2cm, font=\small},
    fast/.style={rectangle, draw=green!60!black, thick, fill=green!5, text width=3.5cm, align=center, rounded corners, minimum height=1.2cm, font=\small},
    data/.style={cylinder, draw=black, thick, aspect=0.25, fill=yellow!10, text width=3.5cm, align=center, minimum height=1.2cm, shape border rotate=90, font=\small},
    phasebox/.style={rectangle, draw=gray, thick, dashed, inner sep=10pt, rounded corners},
    arrow/.style={->, thick, draw=black!80}
]

% --- SHARED INPUT (Top Center) ---
\node [shared] (shared_input) {Shared Candidate Generation\\Target $\mathbf{T}$ \& Point Cloud $\mathcal{P}$};

% --- BASELINE PATH (Left Side) ---
\node [below left=1.5cm and 0.8cm of shared_input.south, font=\bfseries\small] (base_title) {Baseline Planner};
\node [block, below=0.1cm of base_title] (base_orient) {Orientation Sampling\\($M=80$ steps per point)};
\node [bottleneck, below=of base_orient] (base_ik) {Exhaustive Online IK\\(Generate-and-Test)};
\node [below=0.1cm of base_ik, font=\small\color{red!80!black}] (base_time) {\textit{Latency: 13--27\,s}};

% Background Box for Baseline
\begin{scope}[on background layer]
    \node [phasebox, fit=(base_title) (base_orient) (base_time), fill=red!2] (baseline_box) {};
    % \node [above=0.1cm of base_orient.north, font=\bfseries\small] {Baseline Planner};
\end{scope}

% --- PROPOSED ORM PATH (Right Side) ---
\node [below right=1.5cm and 0.8cm of shared_input.south, font=\bfseries\small] (proposed_title) {Proposed ORM Pipeline};
\node [data, below= 0.1cm of proposed_title] (orm_db) {Offline ORM\\(83\,MB)};
\node [fast, below=of orm_db] (orm_lookup) {$\mathcal{O}(1)$ Bitwise Lookup\\(Filter via ORM)};
\node [block, below=of orm_lookup] (orm_verify) {Selective IK Verify\\\& Interpolation};
\node [below=0.1cm of orm_verify, font=\small\color{green!60!black}] (orm_time) {\textit{Latency: 0.289--0.35\,s}};

% Background Box for Proposed
\begin{scope}[on background layer]
    \node [phasebox, fit= (proposed_title) (orm_db) (orm_time), fill=green!2] (proposed_box) {};
    % \node [below=0.1cm of proposed_box.north, font=\bfseries\small] (proposed_title) {Proposed ORM Pipeline};
\end{scope}

% --- OUTPUT (Bottom Center) ---
\path (baseline_box.south) -- (proposed_box.south) coordinate[midway] (box_bottom);
\node [shared, below=2.5cm of box_bottom, text width=4cm] (output) {Feasible Entry Points $\mathcal{E}_{\text{feas}}$};

% --- CONNECTIONS ---
% Shared to Branches
\draw [arrow] (shared_input.south) -- ++(0,-0.4) -| (baseline_box.north);
\draw [arrow] (shared_input.south) -- ++(0,-0.4) -| (proposed_box.north);

% Baseline flow
\draw [arrow] (base_orient) -- (base_ik);
\draw [arrow] (baseline_box.south) -- ++(0,-3.5) -| (output);

% Proposed flow
\draw [arrow] (orm_db) -- (orm_lookup) node[midway, right, font=\footnotesize] {\textit{Loaded once}};
\draw [arrow] (orm_lookup) -- (orm_verify);
\draw [arrow] (proposed_box.south) -- ++(0,-0.5) -| (output);

\end{tikzpicture}%
}
\caption{System architecture of the baseline and proposed ORM-based entry point planning pipelines. The proposed method shifts kinematic feasibility evaluation offline and performs online filtering through indexed reachability lookup.}
\label{fig:architecture}
\end{figure}
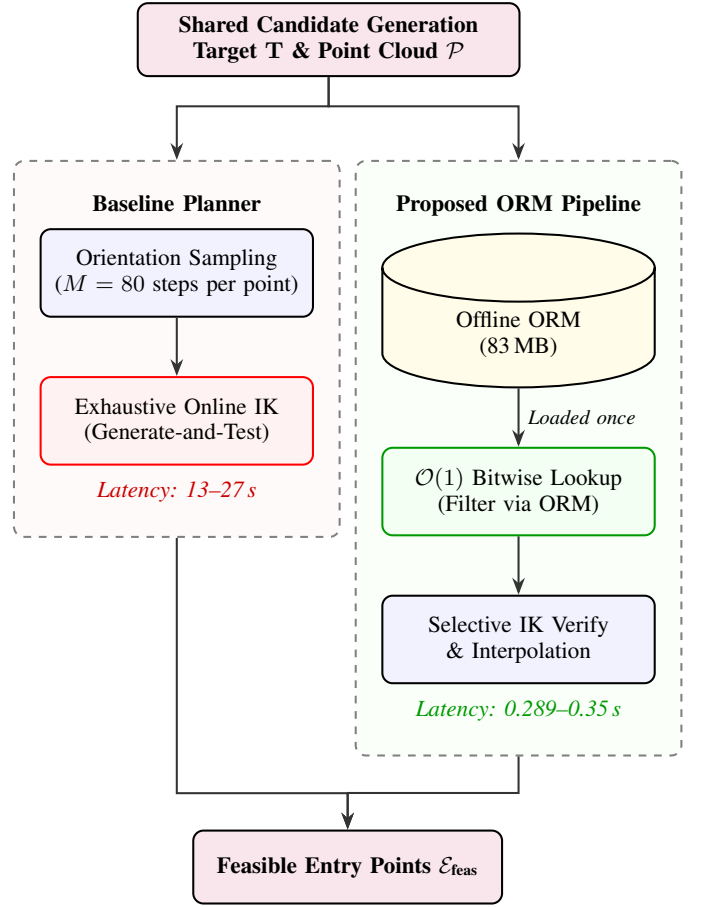

This section describes the proposed reachability-based entry point planning framework. As illustrated in Fig. \ref{fig:architecture}, the system architecture of the proposed pipeline is divided into an offline precomputation phase and an online runtime query phase. We first present the shared candidate generation stage, followed by the baseline validation procedure, the ORM construction, and the runtime planning pipeline.

\subsection{Shared Candidate Entry Point Generation}
\label{subsec:shared_candidate_generation}

Both the baseline and the proposed ORM-based planner operate on an identical set of candidate entry points. The robot workspace is approximated offline by sampling the admissible configuration space $\mathcal{C}$ and mapping each sampled joint configuration $\mathbf{q}_k \in \mathcal{C}$ to Cartesian space via the positional forward kinematics:
\begin{equation}
\mathbf{p}_k = f_p(\mathbf{q}_k).
\end{equation}
The resulting discrete workspace point cloud is defined as
\begin{equation}
\mathcal{P} = \left\{ \mathbf{p}_1, \mathbf{p}_2, \ldots, \mathbf{p}_{N_{\mathrm{pts}}} \right\} \subset \mathbb{R}^3,
\end{equation}
with $N_{\mathrm{pts}} = 188{,}873$ points in the final implementation. 

This shared offline candidate generation ensures a fair comparison between approaches. While the baseline performs exhaustive online orientation sampling and inverse kinematics verification, the proposed method leverages precomputed reachability information from the ORM for rapid filtering over the same candidate set $\mathcal{P}$.

\subsection{Baseline Entry Point Validation}

For a given intraoperative target $\mathbf{T} \in \mathbb{R}^3$, the baseline planner evaluates each candidate entry point $\mathbf{E}_k \in \mathcal{P}$ using a conventional generate-and-test strategy:
\begin{enumerate}
    \item Insertion Depth Constraint: Candidates exceeding the maximum needle insertion length $L = 200$\,mm are discarded:
    \begin{equation}
        \|\mathbf{T} - \mathbf{E}_k\|_2 > L.
    \end{equation}
    \item Orientation Discretization: The normalized insertion direction is computed as
    \begin{equation}
        \mathbf{v}_k = \frac{\mathbf{T} - \mathbf{E}_k}{\|\mathbf{T} - \mathbf{E}_k\|_2},
    \end{equation}
    and the roll angle about this axis is sampled uniformly with $M = 80$ steps over $[0, 2\pi)$.
    \item Feasibility Verification: An analytical inverse kinematics solver is executed for each orientation. A candidate is marked feasible if at least one configuration satisfies joint limits and collision constraints.
\end{enumerate}

This brute-force approach provides the reference for both computational cost and geometric fidelity.

\subsection{Orientation Discretization for ORM Construction}

Accurate representation of the end-effector orientation space $SO(3)$ is essential for constructing a useful reachability map. We parameterize orientations using spherical coordinates for the primary insertion direction, combined with a decoupled roll angle:
\begin{itemize}
    \item Azimuth angle $\phi \in [0, 2\pi)$ and elevation angle $\theta \in [0, \pi]$ define the tool $z$-axis direction.
    \item Roll angle $\gamma \in [0, 2\pi)$ captures rotation about the insertion axis.
\end{itemize}

This $(\phi, \theta, \gamma)$ parameterization is well-suited for needle-based interventions, as it separates the anatomically constrained approach vector from the free instrument roll. The selected discretization resolution balances expressiveness and computational tractability. In our implementation, the orientation space was discretized into $n_\phi = 72$, $n_\theta = 36$, and $n_\gamma = 72$ discrete steps.

This orientation-focused discretization builds upon foundational reachability and capability map techniques \cite{porges-main, dlr88199, junorm}, which typically employ full SE(3) voxel and sphere grids.

\subsection{ORM Construction}

The ORM is constructed offline by exhaustively testing kinematic feasibility across all position-orientation pairs $(\mathbf{p}, (\phi, \theta, \gamma))$. For each pair, an analytical inverse kinematics solver \cite{Sciavicco2000} checks whether a valid joint configuration exists that respects joint limits and self-collision constraints.

To overcome the memory explosion typical of dense 6D reachability maps, we employ a bit-packing scheme. Reachability flags are stored as individual bits in 64-bit unsigned integers (\texttt{uint64\_t}). A 3D orientation index $(i_\phi, i_\theta, i_\gamma)$ is linearized as
\begin{equation}
    k = (i_\phi \cdot n_\theta + i_\theta) \cdot n_\gamma + i_\gamma,
\end{equation}
where $n_\phi$, $n_\theta$, and $n_\gamma$ are the discretization steps along each axis. The word index $w = \lfloor k/64 \rfloor$ and bit offset $b = k \bmod 64$ then determine the storage location.

Unlike general-purpose full SE(3) voxel and sphere grids or hierarchical reductions such as RM4D \cite{rudorfer2024rm4dcombinedreachabilityinverse}, our ORM adopts a task-specific, orientation-focused variant tailored to needle insertion.

\subsection{Runtime Planning Pipeline}

The proposed runtime pipeline replaces exhaustive search with fast bitwise lookups in the precomputed ORM. The full procedure is formalized in Algorithm~\ref{alg:map_based_planning}.

\begin{algorithm}[t]
\caption{ORM-Based Entry Point Planning}
\label{alg:map_based_planning}
\DontPrintSemicolon
\KwIn{Target $\mathbf{T} \in \mathbb{R}^3$, candidate set $\{\mathbf{E}_k\} \subset \mathcal{P}$, precomputed ORM}
\KwOut{Feasible entry points $\mathcal{E}_{\text{feas}}$}
\BlankLine
$\mathcal{E}_{\text{feas}} \leftarrow \emptyset$\;
Load ORM into memory\;
\ForEach{$\mathbf{E}_k \in \{\mathbf{E}_k\}$}{
    Compute insertion direction $\mathbf{v}_k = (\mathbf{T} - \mathbf{E}_k) / \|\mathbf{T} - \mathbf{E}_k\|_2$\;
    % Map $\mathbf{v}_k$ to nearest discrete indices $(i_\phi, i_\theta, i_\gamma)$\;
    Convert $\mathbf{v}_k$ to spherical coordinates and assign center-based nearest-bin indices $(i_\phi, i_\theta, i_\gamma)$\;
    Compute $k_{\text{map}} = (i_\phi \cdot n_\theta + i_\theta) \cdot n_\gamma + i_\gamma$\;
    
    $w \leftarrow \lfloor k_{\text{map}}/64 \rfloor$, \ $b \leftarrow k_{\text{map}} \bmod 64$\;
    \If{($\text{Word}[w] \ \& \ (1 \ll b)) \neq 0$}{
        $\mathcal{E}_{\text{feas}} \leftarrow \mathcal{E}_{\text{feas}} \cup \{\mathbf{E}_k\}$\;
    }
}
Apply linear interpolation along x, y and z at boundaries for smoothing\;
Perform final collision checking and selective IK verification on survivors\;
\Return $\mathcal{E}_{\text{feas}}$\;
\end{algorithm}

By shifting heavy computation offline and using direct bitwise operations, the pipeline transitions from a generate-and-test to a filter-and-verify paradigm. Following the $\mathcal{O}(1)$ bitwise filtering, the surviving candidate set undergoes an interpolation step for boundary smoothing, followed by a selective, lightweight IK verification to guarantee kinematic validity. This enables sub-second planning while preserving geometric fidelity.

\section{Experimental Evaluation}

\subsection{Experimental Setup and Evaluation Metrics}

All experiments were performed in numerical simulation on an Intel Core i7-11800H workstation with 16\,GB RAM using MATLAB R2025a and Python 3.14. Experiments used a simulated 6-DoF serial manipulator from Precision Medtech Co., Ltd. Due to confidentiality obligations under an industry collaboration agreement, the full robot model, joint limits, collision geometry, and analytical IK implementation cannot be disclosed. Nevertheless, all experiments were conducted under identical conditions for both the baseline and ORM-based pipelines.

The experimental protocol compares the ORM-filtered feasible entry point cloud against the baseline cloud. To quantify geometric fidelity, we report the convex hull volume ratio, alpha-shape volume ratio, Hausdorff distance, and Chamfer distance. The alpha-shape volume ratio serves as the primary metric because it better preserves workspace concavities than a convex hull representation, while Hausdorff and Chamfer distances complement the volumetric measures by capturing worst-case and average pointwise deviations between the baseline and ORM-filtered clouds. Together, these metrics provide a balanced assessment of global workspace preservation and local geometric distortion.

\subsection{ORM Construction Results}

The ORM was constructed offline over the shared workspace of
$N_{\mathrm{pts}} = 188{,}873$ discrete positions. To study the trade-off between
orientation resolution, storage cost, and initialization overhead, four discretization profiles were evaluated.

Table~\ref{tab:orm_construction} summarizes the offline construction cost for different orientation discretization profiles. The $72\times36\times72$ configuration was selected for subsequent experiments as a practical trade-off between angular resolution, storage cost, and initialization overhead, resulting in a compressed size of 83\,MB and a mean loading time of 8.0\,s.

\begin{table}[t]
\centering
\caption{ORM Construction Metrics for Different Orientation Discretization Profiles}
\label{tab:orm_construction}
\setlength{\tabcolsep}{7.5pt}
\begin{tabular}{lcccc}
\toprule
Metric & 
\makecell{72$\times$36\\$\times$72} & 
\makecell{60$\times$30\\$\times$60} & 
\makecell{36$\times$18\\$\times$90} & 
\makecell{72$\times$36\\$\times$18} \\
\midrule
Generation time (s) & 17169.3 & 12062.4 & 2759.98 & 2453.76 \\
Orientations / Point & 186624 & 108000 & 58320 & 46656 \\
Load time (s) & 8.0 & 4.2 & 1.6 & 1.3 \\
Size (MB) & 83.0 & 53.9 & 29.5 & 26.7 \\
\bottomrule
\end{tabular}
\end{table}

\begin{figure}[htbp]
\centering
\includegraphics[width=0.49\textwidth]{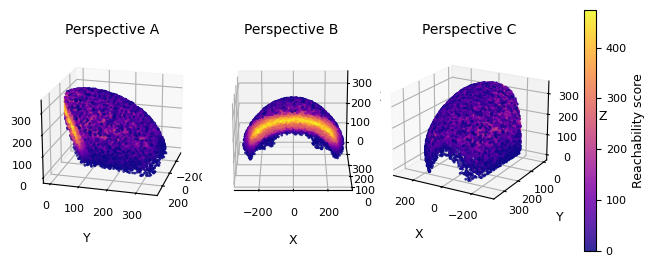}
\caption{Reachability map visualized from three perspectives. Color intensity represents the reachability score (count of feasible orientations) mapped at each workspace coordinate location.}
\label{fig:reachability_map_views}
\end{figure}

Figure~\ref{fig:reachability_map_views} visualizes the selected ORM from three viewpoints. The color intensity denotes the reachability score, i.e., the number of feasible orientations associated with each discrete workspace position.

\subsection{Proposed vs Baseline Comparison}

The proposed ORM-based planner was evaluated against the brute-force baseline across five representative target locations. As shown in Table~\ref{tab:baseline_vs_orm}, the baseline required 13.5--26.9\,s per target due to exhaustive orientation sampling and repeated IK calls. In contrast, the ORM pipeline reduced planning time to 0.289--0.349\,s per target, corresponding to an approximate 60--90$\times$ speedup.

\begin{table*}[t]
\centering
\caption{Performance Comparison: Baseline vs. ORM-based Planning}
\label{tab:baseline_vs_orm}
\setlength{\tabcolsep}{5pt} % Slightly relaxed padding for a clean full-width look
\begin{tabular}{lrrrrrrrr}
\toprule
& \multicolumn{2}{c}{Time} & \multicolumn{2}{c}{Feasible Points} & \multicolumn{2}{c}{Volume Ratios} & \multicolumn{2}{c}{Distance Metrics} \\
\cmidrule(lr){2-3} \cmidrule(lr){4-5} \cmidrule(lr){6-7} \cmidrule(lr){8-9}
Target $(X,Y,Z)$ (mm) & Baseline (s) & ORM (s) & Baseline \# & ORM \# & Convex Hull & Alpha Shape & Hausdorff P95 (mm) & Chamfer (mm) \\

\midrule
$(-90.091,309.047,258.281)$ & 17.702 & 0.349 & 12,988 & 11,955 & 0.940 & 1.151 & 10.250 & 0.325 \\
$(-37.371,258.235,273.095)$ & 26.851 & 0.332 & 11,697 & 10,534 & 0.911 & 0.984 & 10.417 & 0.404 \\
$(9.948,276.207,246.012)$   & 23.793 & 0.300 & 6,431  & 5,956  & 0.920 & 1.067 & 8.514  & 0.494 \\
$(-55.242,288.334,253.022)$ & 22.940 & 0.342 & 10,187 & 9,335  & 0.920 & 0.967 & 9.913  & 0.394 \\
$(11.977,339.075,276.779)$  & 13.481 & 0.289 & 10,130 & 9,926  & 0.938 & 0.849 & 8.466  & 0.297 \\
\bottomrule
\end{tabular}
\end{table*}

\begin{figure*}[htbp]
\centering
\includegraphics[width=\textwidth]{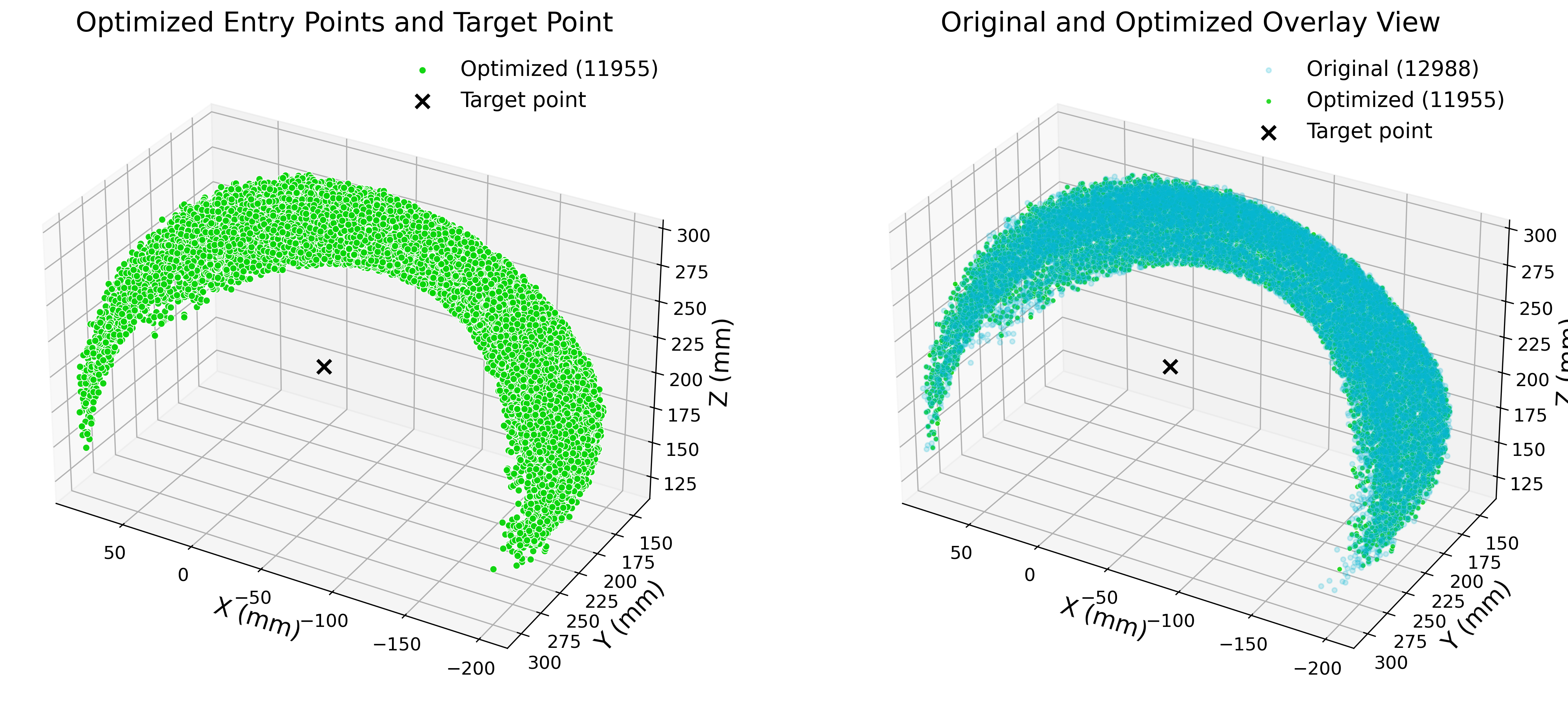}
\caption{Entry point planning result for target $\mathbf{T}_1 = (-90.091, 309.047, 258.281)$\,mm. Left: ORM-filtered feasible points (green). Right: Overlap between baseline (blue) and ORM (green) clouds.}
\label{fig:target_point_result}
\end{figure*}

Figure~\ref{fig:target_point_result} illustrates the spatial distribution of the ORM-derived feasible entry point cloud ($\mathcal{E}_{\text{feas}}$) for the representative target, together with its point-wise overlap against the brute-force baseline. The optimized cloud remains closely aligned with the baseline support while requiring only a fraction of the online computation time.

Despite the large reduction in online computation time, the ORM-based planner preserved the global structure of the feasible entry point cloud with high fidelity. The convex hull volume ratio ranged from 0.891 to 0.955, with a mean of 0.933, and the alpha-shape volume ratio ranged from 0.849 to 1.151, with a mean of 0.976. Boundary deviations remained limited, with P95 Hausdorff distances between 8.514\,mm and 10.417\,mm and Chamfer distances between 0.297\,mm and 0.494\,mm.

\begin{figure}[htbp]
\centering
\includegraphics[width=0.5\textwidth]{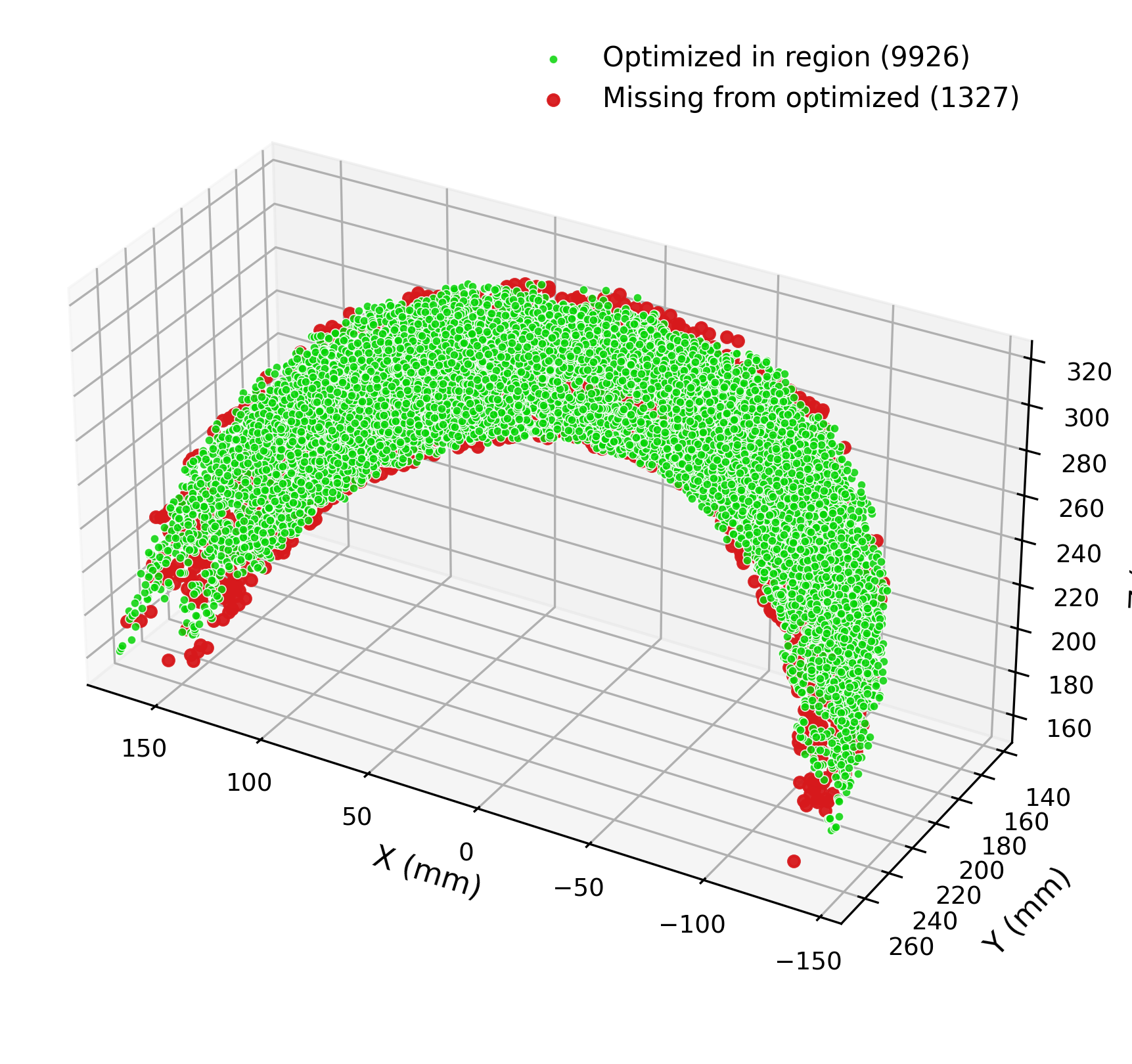}
\caption{Zoomed boundary view of the reconstructed entry point cloud for target point $\mathbf{T}_5 = (11.977, 339.075, 276.779)\,\text{mm}$, showing optimized regions (green) and locally omitted boundary subsets (red).}
\label{fig:zoomed_region}
\end{figure}

Localized boundary discrepancies are illustrated in Fig.~\ref{fig:zoomed_region} for the representative target $\mathbf{T}_5 = (11.977, 339.075, 276.779)\,\text{mm}$. These discrepancies appear as small omitted subsets near peripheral cloud boundaries and are consistent with discretization effects introduced by the selected $72\times36\times72$ orientation grid, while the overall reachable topology remains well preserved.

In a larger-scale evaluation consisting of 10 trials $\times$ 32 targets, the ORM pipeline achieved a mean planning time of $0.188 \pm 0.001$\,s per target (total batch time $6.004 \pm 0.044$\,s). This faster amortized performance compared to the per-target measurements is due to loading the ORM once per batch and benefiting from vectorized lookups across multiple targets. The dominant remaining costs were map lookup, final IK verification on survivors, and boundary smoothing. These results demonstrate that the proposed approach successfully eliminates the computational bottleneck of online orientation search while maintaining strong geometric agreement.

\section{Conclusion and Future Work}

This paper presented an Orientation-Based Reachability Map (ORM) framework to reduce the computational burden of robotic entry point planning for percutaneous liver tumor ablation. By precomputing kinematic feasibility over 188,873 workspace positions and storing orientation reachability in a compact 83\,MB bit-packed representation, the proposed method replaces exhaustive online inverse kinematics evaluation with indexed lookup.

Experimental results demonstrate that the ORM-based pipeline reduces planning latency from 13.5--26.9\,s (baseline) to 0.289--0.349\,s per target while preserving high geometric fidelity, with a mean alpha-shape volume ratio of 0.976 and Chamfer distances below 0.5\,mm. In large-scale multi-target evaluations, the system achieved a mean planning time of $0.188 \pm 0.001$\,s per target, representing a 60--90$\times$ speedup. These findings demonstrate that task-specific orientation-based reachability maps provide an effective basis for low-latency, workspace-aware entry-point planning in robotic percutaneous liver interventions.

Ultimately, this framework makes two primary contributions: a practical methodology for constructing and compressing high-resolution 6D reachability maps, and a JIT-compiled runtime pipeline that shifts the planning paradigm from a generate-and-test approach to an efficient filter-and-verify model. Together, these advancements establish that precomputed orientation-based reachability maps offer a practical and effective solution for real-time robotic needle guidance in time-critical interventional procedures.

\subsection{Future Work}

Future extensions will focus on enhancing the clinical readiness and performance of the framework. First, we plan to conduct a direct comparative evaluation against state-of-the-art reachability-map formulations to quantify the advantages and limitations of the proposed ORM in terms of compression, lookup efficiency, and approximation quality. Second, we will explore continuous reachability representations using deep manifold learning techniques (e.g., variational autoencoders, diffusion models) to eliminate discretization-induced boundary artifacts. Third, the current filter-and-verify pipeline will be extended with manipulability-aware ranking and collision-aware trajectory optimization to select not only feasible but also high-quality insertion paths. Finally, we will integrate dynamic updates to handle minor intraoperative changes and progress toward hardware-in-the-loop validation using anatomical phantoms, followed by ex-vivo and in-vivo studies.

By addressing the fundamental latency barrier in workspace-aware planning, this work represents a significant step toward safe, responsive, and clinically viable robotic assistance in percutaneous liver interventions.

\section*{Acknowledgment}

The authors would like to thank Precision Medtech Co., Ltd., and the Pattern Recognition Lab at Friedrich-Alexander-Universität Erlangen-Nürnberg for their support with the robotic infrastructure and lab environment.

% \bibliographystyle{IEEEtran}
% \bibliography{ref.bib}

% \section*{Data and Code Availability}

% The datasets and source code associated with this work are not publicly available due to confidentiality and intellectual property restrictions under an industry collaboration agreement.

\begin{sloppypar}
\bibliographystyle{IEEEtran}
\bibliography{ref.bib}
\end{sloppypar}

\vspace{12pt}
\color{red}

\end{document}